\pdfoutput=1

\documentclass[11pt]{article}

\usepackage[]{acl}

\usepackage{times}
\usepackage{latexsym}
\usepackage{graphicx}
\usepackage{amsmath}
\usepackage{latexsym}
\usepackage{amssymb}
\usepackage{amsthm}
\usepackage{color}
\usepackage{verbatim}
\usepackage{dsfont}
\usepackage{algorithm}
\usepackage{booktabs}
\usepackage{xspace}
\usepackage[noend]{algpseudocode}
\usepackage[geometry]{ifsym}

\DeclareMathOperator*{\argmax}{arg\,max}

\usepackage[T1]{fontenc}

\usepackage[utf8]{inputenc}

\usepackage{microtype}

\newcommand{\ourmodel}{{\textsc{SelfExplain}}}

\newcommand{\Sref}[1]{\S\ref{#1}}

\author{
  Dheeraj Rajagopal$^\clubsuit$ \quad Vidhisha Balachandran$^\clubsuit$ \quad Eduard Hovy$^\clubsuit$ \quad Yulia Tsvetkov$^\spadesuit$\\
  \vspace{0.5mm}$^\clubsuit$Language Technologies Institute, Carnegie Mellon University \\
$^\spadesuit$Paul G.~Allen School of Computer Science \& Engineering, University of Washington \\
        {\tt \{dheeraj,vbalacha,hovy\}@cs.cmu.edu, yuliats@cs.washington.edu}
}

\begin{document}

\title{\ourmodel: A Self-Explaining Architecture for  Neural Text Classifiers}
\maketitle


\newcommand{\roberta}{RoBERTa}
\newcommand{\xlnet}{XLNet}

\newcommand{\GIL}{{\fontfamily{qcr}\selectfont
GIL\xspace%
}}
\newcommand{\LIL}{{\fontfamily{qcr}\selectfont
LIL\xspace%
}}

\newcommand{\squishlist}{
  \begin{list}{$\bullet$}
    { \setlength{\itemsep}{0pt}      \setlength{\parsep}{3pt}
      \setlength{\topsep}{3pt}       \setlength{\partopsep}{0pt}
      \setlength{\leftmargin}{1.5em} \setlength{\labelwidth}{1em}
      \setlength{\labelsep}{0.5em} } }
\newcommand{\reallysquishlist}{
  \begin{list}{$\bullet$}
    { \setlength{\itemsep}{0.1pt}    \setlength{\parsep}{0.1pt}
      \setlength{\topsep}{0.5pt}     \setlength{\partopsep}{0.5pt}
      \setlength{\leftmargin}{0.5em} \setlength{\labelwidth}{0.5em}
      \setlength{\labelsep}{0.2em} } }

 \newcommand{\squishend}{
     \end{list} 
 }
\begin{abstract}

We introduce \textbf{\ourmodel}, a novel self-explaining model that explains a text classifier's predictions using phrase-based concepts. 
\ourmodel~augments existing neural classifiers by adding (1) a \emph{globally interpretable layer} that identifies the most influential concepts in the training set for a given sample and (2) a \emph{locally interpretable layer} that quantifies the contribution of each local input concept by computing a relevance score relative to the predicted label.
Experiments across five text-classification datasets show that 
\ourmodel~ facilitates interpretability without sacrificing performance. Most importantly, explanations from \ourmodel~show sufficiency for model predictions and are perceived as adequate, trustworthy and understandable by human judges compared to existing widely-used baselines.\footnote{Code and data is publicly available at \url{https://github.com/dheerajrajagopal/SelfExplain}} 
\end{abstract}

\section{Introduction} 

Neural network models are often opaque: they provide limited insight into interpretations of model decisions and are typically treated as ``black boxes'' \citep{lipton2018mythos}. There has been ample evidence that such models overfit to spurious artifacts \citep{gururangan-etal-2018-annotation,McCoy2019RightFT,kumar-etal-2019-topics} and amplify biases in data \citep{zhao2017men,sun-etal-2019-mitigating}. This underscores the need to understand model decision making. 

\begin{figure}[t]
    {\includegraphics[width=\columnwidth]{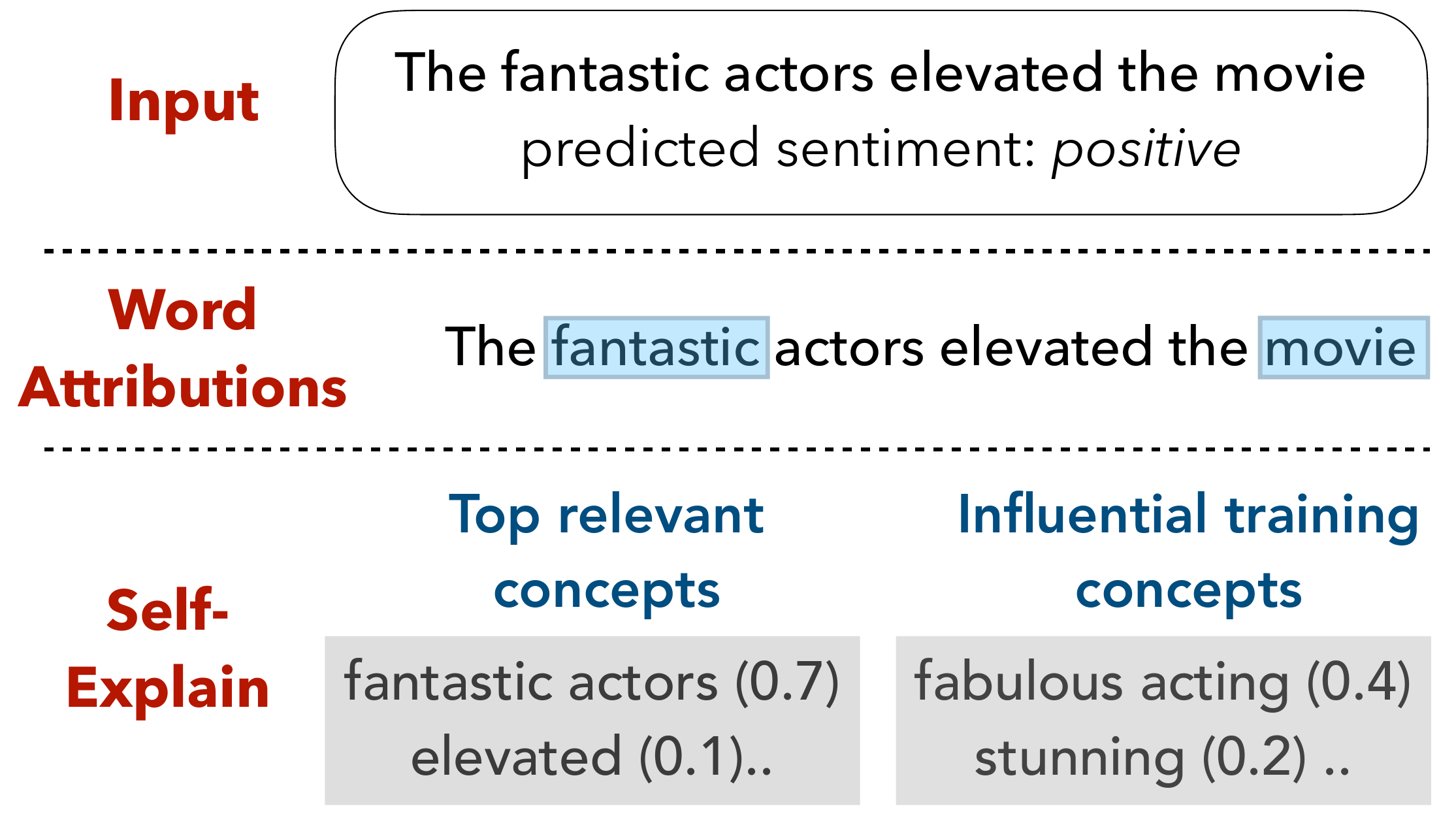}}
    \caption{A sample of interpretable concepts from \ourmodel~ for a binary sentiment analysis task. Compared to saliency-map style word attributions, \ourmodel~ can provide explanations via concepts in the input sample and the concepts in the training data}
    \label{fig:local_global_example}
\end{figure}


Prior work in interpretability for neural text classification predominantly follows two approaches: (i) \emph{post-hoc explanation methods} that explain predictions for previously trained models based on model internals, and (ii) \emph{inherently interpretable models} whose interpretability is built-in and  optimized jointly with the end task. While post-hoc methods \citep{Simonyan2014DeepIC,koh2017understanding,ribeiro-etal-2016-trust} are often the only option for already-trained models, inherently interpretable models \citep{melis2018towards,Arik2020ProtoAttendAP} may provide greater transparency since explanation capability is embedded directly within the model \citep{kim2014bayesian,doshi2017towards,rudin2019stop}. 

In natural language applications, feature attribution based on attention scores \citep{Xu2015ShowAA} has been the predominant method for developing inherently interpretable neural classifiers. Such methods interpret model decisions \emph{locally} by explaining the classifier's decision as a function of relevance of features (words) in input samples. 
However, such interpretations were shown to be unreliable \citep{serrano-smith-2019-attention,pruthi2019learning} and unfaithful \citep{jain-wallace-2019-attention,wiegreffe-pinter-2019-attention}. 
Moreover, with natural language being structured and compositional, explaining the role of higher-level compositional concepts like phrasal structures (beyond individual word-level feature attributions) remains an open challenge. 
Another known limitation of such feature attribution based methods is that the explanations are limited to the input feature space and often require additional methods \citep[e.g.][]{han2020explaining} for providing global explanations, i.e., explaining model decisions as a function of influential training data. 

In this work, we propose \ourmodel---a self explaining model that incorporates both global and local interpretability layers into neural text classifiers. Compared to word-level feature attributions, we use high-level phrase-based concepts, producing a more holistic picture of a classifier's decisions. 
\ourmodel~incorporates: 
(i) \emph{Locally Interpretable Layer} (\LIL), a layer that quantifies via activation difference, the relevance of each concept to the final label distribution of an input sample. 
(ii) \emph{Globally Interpretable Layer} (\GIL), a layer that uses maximum inner product search (MIPS) to retrieve the most influential concepts from the training data for a given input sample. 
We show how \GIL~and \LIL~layers can be integrated into transformer-based classifiers, converting them into self-explaining architectures.  The interpretability of the classifier is enforced through regularization \citep{melis2018towards}, and the entire model is end-to-end differentiable. To the best of our knowledge, \ourmodel~is the first self-explaining neural text classification approach to provide both global and local interpretability in a single model. 

Ultimately, this work makes a step towards combining the generalization power of neural networks with the benefits of interpretable statistical classifiers with hand-engineered features: our experiments on three text classification tasks spanning five datasets with pretrained transformer models show that incorporating \LIL~and \GIL~layers facilitates richer interpretability while maintaining end-task performance. The explanations from \ourmodel~sufficiency reflect model predictions and are perceived by human annotators as more understandable, adequately justifying the model predictions and trustworthy, compared to strong baseline interpretability methods. 

\section{\ourmodel}
\label{sec:model}

Let $\mathcal{M}$ be a neural $C$-class classification model that maps $\mathcal{X} \rightarrow \mathcal{Y}$, where $\mathcal{X}$ are the inputs and $\mathcal{Y}$ are the outputs.  \ourmodel~ builds into $\mathcal{M}$, and it provides a set of explanations $\mathcal{Z}$ via  high-level ``concepts'' that explain the classifier's predictions. We first define interpretable concepts in \Sref{sec:concepts}. We then describe how these concepts are incorporated into a concept-aware encoder in \Sref{sec:concept_aware_encoder}.
In \Sref{sec:lil}, we define our Local Interpretability Layer (\LIL), which provides local explanations by assigning relevance scores to the constituent concepts of the input. In \Sref{sec:gil}, we define our Global Interpretability Layer (\GIL), which provides global explanations by retrieving influential concepts from the training data. Finally, in \Sref{subsec:training}, we describe the end-to-end training procedure and optimization objectives.

\subsection{Defining human-interpretable concepts}
\label{sec:concepts}

Since natural language is highly compositional \citep{montague1970english}, it is essential that interpreting a text sequence goes beyond individual words. We define the set of basic units that are interpretable by humans as \emph{concepts}. In principle, concepts can be words, phrases, sentences, paragraphs or abstract entities. In this work, we focus on phrases as our concepts, specifically all non-terminals in a constituency parse tree. 
Given any sequence $\mathbf{x} = \{w_i\}_{1:T}$, we decompose the sequence into its component non-terminals $N(\mathbf{x}) = \{ nt_j \}_{1:J}$, where $J$ denotes the number of non-terminal phrases in $\mathbf{x}$.

Given an input sample $\mathbf{x}$, $\mathcal{M}$ is trained to produce two types of explanations: 
(i) global explanations from the training data $\mathcal{X}_{train}$ and (ii) local explanations, which are phrases in $\mathbf{x}$. We show an example in Figure~\ref{fig:local_global_example}. 
Global explanations are achieved by identifying the most influential concepts $\mathcal{C}_G$ from the ``concept store'' $\mathbf{Q}$, which is constructed to contain all concepts from the training set $\mathcal{X}_{train}$ by extracting phrases under each non-terminal in a syntax tree for every data sample (detailed in \Sref{sec:gil}). 
Local interpretability is achieved by decomposing the input sample $\mathbf{x}$ into its constituent phrases under each non-terminal in its syntax tree. Then each concept is assigned a score that quantifies its contribution to the sample's label distribution for a given task; $\mathcal{M}$ then outputs the most relevant local concepts $\mathcal{C}_L$.

\begin{figure*}[!ht]
    \centering
    {\includegraphics[width=0.95\textwidth]{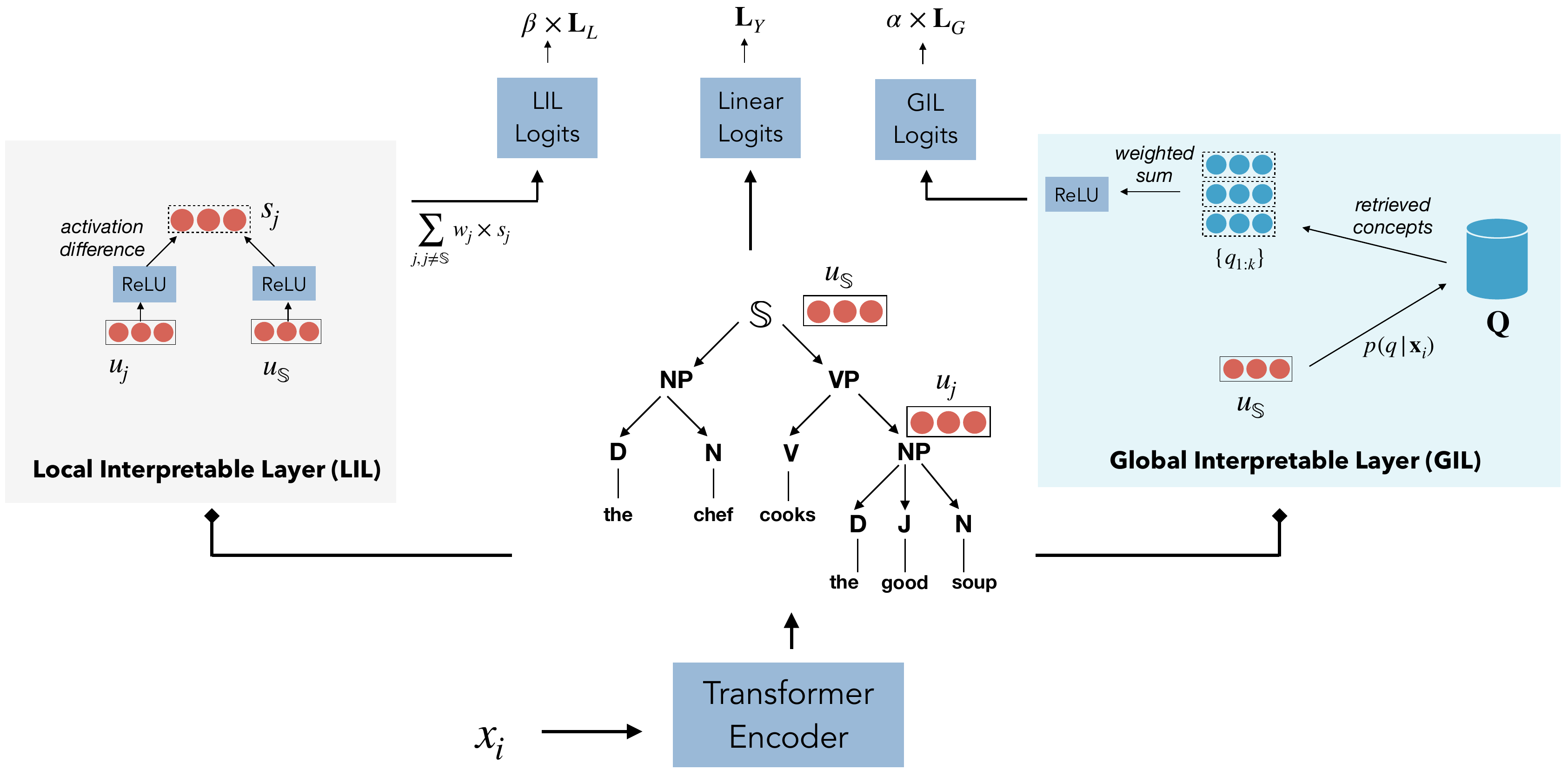}}
    \caption{Model Architecture: Our architecture comprises a base encoder that encodes the input and its relative non-terminals. \GIL~ then uses MIPS to retrieve the most influential concepts  that \emph{globally} explain the sample, while \LIL~ computes a relevance score for each $nt_j$ that quantifies its relevance to predict the label. The model interpretability is enforced through regularization. Examples of top \LIL~concepts (extracted from the from input) are \{\emph{the good soup}, \emph{good}\}, and of top \GIL~concepts (from the training data) are \{\textit{great food, excellent taste}\}}
    \label{fig:architecture}
\end{figure*}

\subsection{Concept-Aware Encoder $\mathbf{E}$}
\label{sec:concept_aware_encoder}

We obtain the encoded representation of our input sequence $\mathbf{x} = \{w_i\}_{1:T}$ from a pretrained transformer model \cite{vaswani2017attention, liu2019roberta, yang2019xlnet} by extracting the final layer output as $\{\mathbf{h}_i\}_{1:T}$.
Additionally, we compute representations of concepts, $\{\mathbf{u}_j\}_{1:J}$. For each non-terminal $nt_j$ in $\mathbf{x}$, we represent it as the mean of its constituent word representations $\mathbf{u}_j = \dfrac{\sum_{w_i \in nt_j} \mathbf{h}_i}{len(nt_j)}$ where $len(nt_j)$ represents the number of words in the phrase $nt_j$. To represent the root node ($\mathbb{S}$) of the syntax tree, $nt_{\mathbb{S}}$, we use the pooled representation (\texttt{[CLS]} token representation) of the pretrained transformer as $\mathbf{u}_{\mathbb{S}}$ for brevity.\footnote{We experimented with different pooling strategies (mean pooling, sum pooling and pooled \texttt{[CLS]} token representation) and all of them performed similarly. We chose to use the pooled \texttt{[CLS]} token for the final model as this is the most commonly used method for representing the entire input.} 
Following traditional neural classifier setup, the output of the classification layer $l_Y$ is computed as follows:
\begin{align}
    l_Y &= \texttt{softmax}(\mathbf{W}_y \times g(\mathbf{u}_{\mathbb{S}}) + \mathbf{b}_y) \nonumber \\
    P_C &= \argmax(l_Y) \nonumber
\end{align}
where $g$ is a $relu$ activation layer, $\mathbf{W}_y \in \mathbb{R}^{D \times C}$, and $P_C$ denotes the index of the predicted class. 

\subsection{Local Interpretability Layer (\textbf{\LIL})}
\label{sec:lil}
For local interpretability, we compute a local relevance score for all input concepts  $\{nt_j\}_{1:J}$ from the sample $\mathbf{x}$. Approaches that assign relative importance scores to input features through activation differences \citep{pmlr-v70-shrikumar17a,Montavon2017ExplainingNC} are widely adopted for interpretability in computer vision applications. Motivated by this, we adopt a similar approach to NLP applications where we learn the attribution of each concept to the final label distribution via their activation differences. Each non-terminal $nt_j$ is assigned a score that quantifies the contribution of each $nt_j$ to the label in comparison to the contribution of the root node $nt_{\mathbb{S}}$. The most contributing phrases $\mathcal{C}_L$ is used to locally explain the model decisions. 

Given the encoder $\mathbf{E}$, \LIL~ computes the contribution solely from $nt_j$ to the final prediction. We first build a representation of the input without contribution of phrase $nt_j$ and use it to score the labels:
\vspace{-7mm}
\begin{align}
t_j &=  g(\mathbf{u}_j) - g(\mathbf{u}_{\mathbb{S}}) \nonumber \\
s_j &= \texttt{softmax}(\mathbf{W}_v \times t_j + \mathbf{b}_v)  \nonumber
\end{align} 
where $g$ is a $relu$ activation function, $t_j \in \mathbb{R}^D$, $s_j \in \mathbb{R}^C$,  $\mathbf{W}_v \in \mathbb{R}^{D \times C} $. 
Here, $s_j$ signifies a label distribution without the contribution of $nt_j$. Using this, the relevance score of each $nt_j$ for the final prediction is given by the difference between the classifier score for the predicted label based on the entire input and the label score based on the input without $nt_j$:
$\mathbf{r}_j = (l_{Y})_i\rvert_{i = P_C} - (s_j)_i\rvert_{i = P_C} \nonumber, $
where $\mathbf{r}_j$ is the relevance score of the concept $nt_j$.

\subsection{Global Interpretability layer (\textbf{\GIL})} 
\label{sec:gil}

The Global Interpretability Layer \GIL~aims to interpret each data sample $\mathbf{x}$ by providing a set of $K$ concepts from the training data which most influenced the model's predictions. Such an approach is advantageous as we can now understand how important concepts from the training set influenced the model decision to predict the label of a new input, providing more granularity than methods that use entire samples from the training data for post-hoc interpretability \citep{koh2017understanding,han2020explaining}. 

We first build a \emph{concept store} $Q$ which holds all the concepts from the training data. Given model $\mathcal{M}$ , we represent each concept candidate from the training data, $q_k$ as a mean pooled representation of its constituent words $q_k = \dfrac{\sum_{w \in q_k} e(w)}{len(q_k)} \in \mathbb{R}^D$, where $e$ represents the embedding layer of $\mathcal{M}$ and $len(q_k)$ represents the number of words in $q_k$.  
$Q$ is represented by a set of $\{q\}_{1:N_Q}$, which are $N_Q$ number of concepts from the training data. As the model $\mathcal{M}$ is finetuned for a downstream task, the representations $q_k$ are constantly updated. Typically, we re-index all candidate representations $q_k$ after every fixed number of training steps.

For any input $\mathbf{x}$, \GIL~produces a set of $K$ concepts $\{q\}_{1:K}$ from $Q$ that are most influential as defined by the cosine similarity function:
$$d(\mathbf{x}, Q) = \dfrac{\mathbf{x} \cdot q}{\| \mathbf{x} \| \| q \|} \quad \forall q \in Q $$

Taking $\mathbf{u}_{\mathbb{S}}$ as input, \GIL~uses dense inner product search to retrieve the top-$K$ influential concepts $\mathcal{C}_G$ for the sample. Differentiable approaches through Maximum Inner Product Search (MIPS) has been shown to be effective in Question-Answering settings \citep{guu2020realm, Dhingra2020Differentiable} to leverage retrieved knowledge for reasoning \footnote{MIPS can often be efficiently scaled using approximate algorithms \citep{shrivastava2014asymmetric} }. Motivated by this, we repurpose this retrieval approach to identify the influential concepts from the training data and learn it end-to-end via backpropagation. Our inner product model for \GIL~is defined as follows:

$$p (q | \mathbf{x}_i) = \dfrac{exp \; d(\mathbf{u}_{\mathbb{S}}, q)}{\sum_{q'}exp \; d(\mathbf{u}_{\mathbb{S}}, q')} \nonumber$$

\subsection{Training}
\label{subsec:training}
\ourmodel~is trained to maximize the conditional log-likelihood of predicting the class at all the final layers: linear (for label prediction), \LIL~, and \GIL~. Regularizing models with explanation specific losses have been shown to improve inherently interpretable models \citep{melis2018towards} for local interpretability. We extend this idea for both global and local interpretable output for our classifier model. 
For our training, we regularize the loss through \GIL~and \LIL~layers by optimizing their output for the end-task as well. 
For the \GIL~layer, we aggregate the scores over all the retrieved $q_{1:K}$ as a weighted sum, followed by an activation layer, linear layer and softmax to compute the log-likelihood loss as follows: 
\begin{align}
    l_{G} &= \texttt{softmax}( \mathbf{W}_u \times g(\sum_{k=1}^K \mathbf{w}_k \times q_k) + \mathbf{b}_u ) \nonumber 
\end{align}
and
$
    \mathcal{L}_G = - \sum_{c=1}^{C} y_c \text{ log}(l_G) 
$
where the global interpretable concepts are denoted by $\mathcal{C}_G = q_{1:K}$, $\mathbf{W}_u \in \mathbb{R}^{D \times C}$, $\mathbf{w}_k \in \mathbb{R}$ and $g$ represents $relu$ activation, and $l_G$ represents the softmax for the \GIL~layer. 

For the \LIL~layer, we compute a weighted aggregated representation over $s_j$ and compute the log-likelihood loss as follows: 
\begin{align}
    l_{L} &= \sum_{j, j \neq \mathbb{S}} \mathbf{w}_{sj} \times s_j,\text{ } \mathbf{w}_{sj}  \in \mathbb{R} \nonumber
\end{align}
and
$\mathcal{L}_L = - \sum_{c=1}^C y_c \text{ log} (l_L) $.
To train the model, we optimize for the following joint loss, $$\mathcal{L} = \alpha \times \mathcal{L}_G + \beta \times \mathcal{L}_L + \mathcal{L}_Y $$ where 
$\mathcal{L}_Y = - \sum_{c=1}^{C} y_c \text{ } log(l_Y)$. 
Here, $\alpha$ and $\beta$ are regularization hyper-parameters.
All loss components use cross-entropy loss based on task label $y_c$.

\section{Dataset and Experiments}

\begin{table}[!ht]
\centering
\begin{tabular}{@{}lrrrr@{}}
\toprule
\textbf{Dataset} & $\textbf{C}$ & $\textbf{L}$ & \textbf{Train} & \textbf{Test} \\ \midrule
SST-2 & 2 & 19 & 68,222 & 1,821 \\
SST-5 & 5 & 18  & 10,754 & 1,101 \\
TREC-6 & 6 & 10  & 5,451 & 500 \\
TREC-50 & 50 & 10  & 5,451 & 499 \\
SUBJ & 2 & 23  & 8,000 & 1,000 \\ \bottomrule
\end{tabular}%
\caption{Dataset statistics, where $\mathbf{C}$ is the number of classes and $\mathbf{L}$ is the average sentence length}
\label{tab:data_stats}
\end{table}

\begin{table*}[!htbp]
\centering
\begin{tabular}{@{}cccccc@{}}
\toprule
         Model    & SST-2 & SST-5 & TREC-6 & TREC-50 & SUBJ\\ \midrule
\xlnet   & 93.4  & 53.8  & \textbf{96.6}  & 82.8 & 96.2   \\
\ourmodel-\xlnet~($K$=5)   & \textbf{94.6}  & \textbf{55.2}  & 96.4   & \textbf{83.0} & \textbf{96.4} \\
\ourmodel-\xlnet~($K$=10)   & 94.4  & 55.2  & 96.4   & 82.8   &  96.4 \\ \midrule

\roberta & 94.8  & 53.5  & 97.0   & 89.0  & 96.2  \\
\ourmodel-\roberta~($K$=5) & \textbf{95.1}  & \textbf{54.3}  & \textbf{97.6}   & \textbf{89.4} & \textbf{96.3} \\
\ourmodel-\roberta~ ($K$=10) & 95.1  & 54.1  & 97.6   & 89.2 & 96.3  \\ \bottomrule
\end{tabular}
\caption{Performance comparison of models with and without \GIL~ and \LIL~ layers. All experiments used the same encoder configurations. We use the development set for SST-2 results (test set of SST-2 is part of GLUE benchmark) and test sets for - SST-5, TREC-6, TREC-50 and SUBJ $\alpha,\beta = 0.1$ for all the above settings.}
\label{tab:performance-nums}
\end{table*}

\paragraph{Datasets:}
\label{subsec:datasets}
We evaluate our framework on five classification datasets:
(i) SST-2
\footnote{https://gluebenchmark.com/tasks}
Sentiment Classification task \cite{socher2013recursive}: the task is to predict the sentiment of movie review sentences as a binary classification task.
(ii) SST-5
\footnote{https://nlp.stanford.edu/sentiment/index.html}
: a fine-grained sentiment classification task that uses the same dataset as before, but modifies it into a finer-grained 5-class classification task. 
(iii) TREC-6 
\footnote{https://cogcomp.seas.upenn.edu/Data/QA/QC/}
: a question classification task proposed by \citet{li2002learning}, where each question should be classified into one of 6 question types. 
(iv) TREC-50: a fine-grained version of the same TREC-6 question classification task with 50 classes 
(v) SUBJ: subjective/objective binary classification dataset \citep{pang2005seeing}. The dataset statistics are shown in Table \ref{tab:data_stats}.

\paragraph{Experimental Settings:}

For our \ourmodel~experiments, we consider two transformer encoder configurations as our base models: 
(1) \roberta~encoder \citep{liu2019roberta} --- a robustly optimized version of BERT \cite{devlin-etal-2019-bert}. 
(2) \xlnet~encoder \cite{yang2019xlnet} --- a transformer model based on Transformer-XL \cite{dai2019transformer} architecture.

We incorporate \ourmodel~ into \roberta~and \xlnet, and use the above encoders without the \GIL~and \LIL~layers as the baselines. We generate parse trees \citep{Kitaev2018ConstituencyPW} to extract target concepts for the input and follow same pre-processing steps as the original encoder configurations for the rest. We also maintain the hyperparameters and weights from the pre-training of the encoders. 
The architecture with \GIL~ and \LIL~ modules are fine-tuned on datasets described in \S \ref{subsec:datasets}. 
For the number of global influential concepts $K$, we consider two settings $K=5,10$. We also perform hyperparameter tuning on $\alpha, \beta = \{ 0.01, 0.1, 0.5, 1.0 \}$ and report results on the  best model configuration. All models were trained on an NVIDIA V-100 GPU.


\paragraph{Classification Results :}
\label{subsec:class_results}


We first evaluate the utility of classification models  after incorporating \GIL~and \LIL~layers in Table \ref{tab:performance-nums}.
%
Across the different classification tasks, we observe that \ourmodel-\roberta~ and \ourmodel-\xlnet~consistently show competitive performance compared to the base models except for a marginal drop in TREC-6 dataset for \ourmodel-\xlnet. 

We also observe that the hyperparameter $K$ did not make noticeable difference. Additional ablation experiments in Table~\ref{tab:ablation} suggest that gains through \GIL~ and \LIL~ are complementary and both layers contribute to performance gains. 

\begin{table}[ht]
\centering
\small
\begin{tabular}{@{}lc@{}}
\toprule
Model & Accuracy \\ \midrule
\xlnet-Base & 93.4 \\
\ourmodel-\xlnet~ + \LIL & 94.3 \\
\ourmodel-\xlnet~ + \GIL & 94.0 \\
\ourmodel-\xlnet~ + \GIL~ + \LIL & 94.6 \\ \midrule
\roberta-Base  & 94.8 \\
\ourmodel-\roberta~ + \LIL & 94.8 \\
\ourmodel-\roberta~ + \GIL & 94.8 \\
\ourmodel-\roberta~ + \GIL~ + \LIL & 95.1 \\ \bottomrule
\end{tabular}
\caption{Ablation: \ourmodel-\xlnet~ and \ourmodel-\roberta~ base models on SST-2.}
\label{tab:ablation}
\end{table}

\section{Explanation Evaluation}
Explanations are notoriously difficult to evaluate quantitatively \citep{DoshiVelez2017AccountabilityOA}.
A \emph{good} model explanation should be (i) relevant to the current input and predictions and (ii) understandable to humans  \citep{deyoung-etal-2020-eraser,jacovi-goldberg-2020-towards,Wiegreffe2020MeasuringAB,jain-etal-2020-learning}. 
Towards this, we evaluate whether the explanations along the following diverse criteria:
\squishlist
    \item \textbf{Sufficiency} -- Do explanations sufficiently reflect the model predictions?   
    \item \textbf{Plausibility} -- Do explanations appear plausible and understandable to humans?
    \item \textbf{Trustability} -- Do explanations improve human trust in model predictions? 
\squishend

From \ourmodel, we extracted (i) \textit{Most relevant local concepts}: these are the top ranked phrases based on $\mathbf{r}(nt)_{1:J}$ from the \LIL~ layer and
(ii) \textit{Top influential global concepts:} these are the most influential concepts $q_{1:K}$ ranked by the output of \GIL~ layer as the model explanations to be used for evaluations.

\subsection{Do \ourmodel~ explanations reflect predicted labels?}
\emph{Sufficiency} aims to evaluate whether model  explanations alone are highly indicative of the predicted label \citep{Jacovi2018UnderstandingCN,Yu2019RethinkingCR}. 
``Faithfulness-by-construction'' (FRESH) pipeline \citep{jain-etal-2020-learning} is an example of such framework to evaluate sufficiency of explanations: the sole explanations, without the remaining parts of the input, must be sufficient for predicting a label. 
In FRESH, a BERT \citep{devlin-etal-2019-bert} based classifier is trained to perform a task using only the extracted explanations without the rest of the input.
An explanation that achieves high accuracy using this classifier is indicative of its ability to recover the original model prediction. 

We evaluate the explanations on the sentiment analysis task. Explanations from \ourmodel~are incorporated to the FRESH framework and we compare the predictive accuracy of the explanations in comparison to baseline explanation methods. 
Following \citet{jain-etal-2020-learning}, we use the same experimental setup and saliency-based baselines such as attention \citep{lei-etal-2016-rationalizing,bastings-etal-2019-interpretable} and gradient \citep{Li2016VisualizingAU} based explanation methods. From Table \ref{tab:quant_eval}\footnote{In these experiments, explanations are pruned at a maximum of 20\% of input. For \ourmodel, we select upto top-$K$ concepts thresholding at 20\% of input}, we observe that \ourmodel~ explanations from \LIL~and \GIL~show high predictive performance compared to all the baseline methods.
Additionally, \GIL~explanations outperform full-text (an explanation that uses all of the input sample) performance, which is often considered an upper-bound for span-based explanation approaches. We hypothesize that this is because \GIL~explanation concepts from the training data are very relevant to help disambiguate the input text. In summary, outputs from \ourmodel~are more predictive of the label compared to prior explanation methods indicating higher sufficiency of explanations. 

\begin{table}[!ht]
\centering
\normalsize
\begin{tabular}{@{}llc@{}}
\toprule
Model & Explanation & Accuracy \\ \midrule
Full input text & - & 0.90 \\
\citet{lei-etal-2016-rationalizing} & \begin{tabular}[c]{@{}l@{}}contiguous\\ top-$K$ tokens \end{tabular} & \begin{tabular}[c]{@{}l@{}}0.71\\ 0.74\end{tabular} \\
\citet{bastings-etal-2019-interpretable} & \begin{tabular}[c]{@{}l@{}}contiguous\\ top-$K$ tokens \end{tabular} & \begin{tabular}[c]{@{}l@{}}0.60\\ 0.59\end{tabular} \\
\citet{Li2016VisualizingAU} & \begin{tabular}[c]{@{}l@{}}contiguous\\ top-$K$ tokens \end{tabular} & \begin{tabular}[c]{@{}l@{}}0.70\\ 0.68\end{tabular} \\
\texttt{[CLS]} Attn & \begin{tabular}[c]{@{}l@{}}contiguous\\ top-$K$ tokens  \end{tabular} & \begin{tabular}[c]{@{}l@{}}0.81\\ 0.81\end{tabular} \\ \midrule
\ourmodel-\LIL & top-$K$ concepts  & \textbf{0.84} \\ 
\ourmodel-\GIL & top-$K$ concepts & \textbf{0.93} \\ \bottomrule
\end{tabular}
\caption{Model predictive performances (prediction accuracy) on SST-dataset test set. 
 \emph{Contiguous} refers to explanations that are spans of text and top-$K$ refers to model-ranked top-$K$ tokens. \ourmodel~ also uses at most top-$K$ (where $K$=2) concepts for both \LIL~and \GIL.
 \ourmodel~ explanations from both \GIL~  and \LIL~ outperform all baselines.}
\label{tab:quant_eval}
\end{table}

\subsection{Are \ourmodel~ explanations plausible and trustable for humans?}

\begin{table*}[!ht]
\centering
\resizebox{\textwidth}{!}{%
\begin{tabular}{llll}
\hline
Sample & $P_C$ & Top relevant phrases from LIL & Top influential concepts from \GIL \\ \hline
\begin{tabular}[c]{@{}l@{}}the iditarod lasts for days -\\ this just felt like it did .\end{tabular} & neg & for days & \begin{tabular}[c]{@{}l@{}}exploitation piece,\\ heart attack\end{tabular} \\ \midrule
\begin{tabular}[c]{@{}l@{}}corny,  schmaltzy and predictable, but still \\ manages to be kind of heart warming, nonetheless.\end{tabular} & pos & corny, schmaltzy, of heart & \begin{tabular}[c]{@{}l@{}}successfully blended satire, \\ spell binding fun\end{tabular} \\ \midrule
\begin{tabular}[c]{@{}l@{}}suffers from the lack of a \\ compelling or comprehensible narrative  .\end{tabular} & neg & comprehensible, the lack of & \begin{tabular}[c]{@{}l@{}}empty theatres,\\ tumble weed\end{tabular} \\ \midrule
\begin{tabular}[c]{@{}l@{}}the structure the film takes may find matt damon \\ and ben affleck once again looking for residuals\\ as this officially completes a \\ good will hunting trilogy that was never planned  .\end{tabular} & pos & the structure of the film & \begin{tabular}[c]{@{}l@{}}bravo, \\ meaning and consolation\end{tabular} \\ \bottomrule
\end{tabular}%
}
\caption{Sample output from the model and its corresponding local and global interpretable outputs SST-2 ($P_C$ stands for predicted class) (some input text cut for brevity). More qualitative examples in appendix \S\ref{subsec:qual_examples}}
\label{tab:qual_examples}
\end{table*}

Human evaluation is commonly used to evaluate \textit{plausibility} and \textit{trustability}. 
To this end, 14 human judges\footnote{Annotators are graduate students in computer science.}
annotated 50 samples from the SST-2 validation set of sentiment excerpts \citep{socher2013recursive}.
Each judge compared local and global explanations produced by the \ourmodel-\xlnet~ model against two commonly used interpretability methods (i) Influence functions \citep{han2020explaining} for global interpretability and (ii) Saliency detection \citep{Simonyan2014DeepIC} for local interpretability. We follow a setup discussed in \citet{han2020explaining}. 
Each judge was provided the evaluation criteria (detailed next) with a corresponding description. 
The models to be evaluated were anonymized and humans were asked to rate them according to the evaluation criteria alone. 

Following \citet{ehsan2019automated}, we analyse the \emph{plausibility} of explanations which aims to understand how users would perceive such explanations if they were generated by humans.  
We adopt two criteria proposed by \citet{ehsan2019automated}: 

\paragraph{Adequate justification}: 
Adequately justifying the prediction is considered to be an important criteria for acceptance of a model \citep{Davis1989PerceivedUP}. We evaluate the \emph{adequacy} of the explanation by asking human judges: ``Does the explanation adequately justifies the model prediction?'' 
Participants deemed explanations that were irrelevant or incomplete as less adequately justifying the model prediction.
Human judges were shown the following: (i) input, (ii) gold label, (iii) predicted label, and (iv) explanations from baselines and \ourmodel. The models were anonymized and shuffled.

Figure \ref{fig:eval_interpret} (left) shows that
\ourmodel~ achieves a gain of 32\% in perceived adequate justification, providing further evidence that humans perceived \ourmodel~explanations as more plausible compared to the baselines.

\begin{figure}[!ht]
    {\includegraphics[width=\columnwidth]{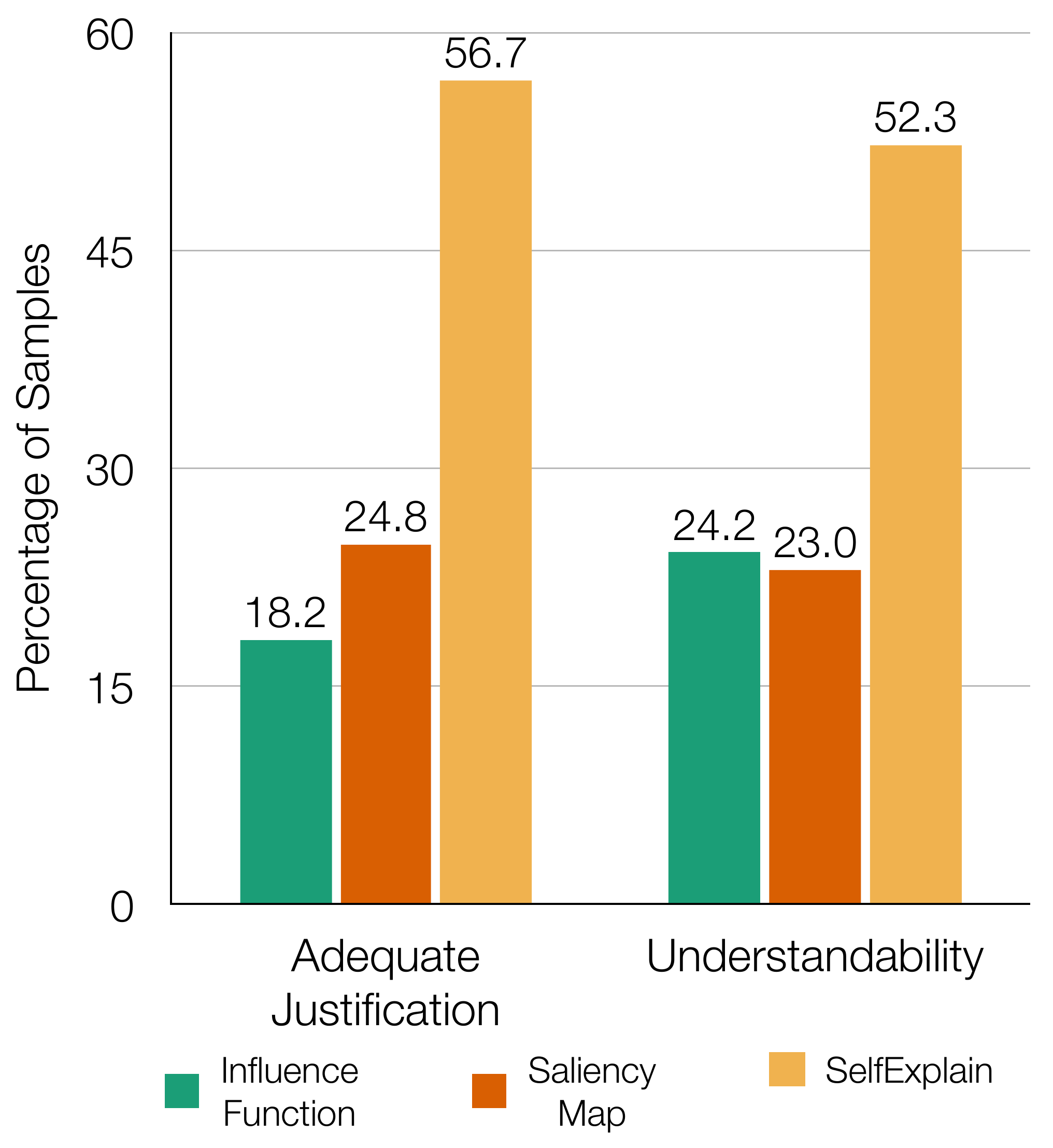}}
    \caption{ \emph{Adequate justification} and \emph{understandability} of \ourmodel~against baselines. The vertical axis shows the percentage of samples evaluated by humans. Humans judge \ourmodel~ explanations to better justify the predictions and be more understandable.}
    \label{fig:eval_interpret}
\end{figure}
\begin{figure}[!h]
    {\includegraphics[width=\columnwidth]{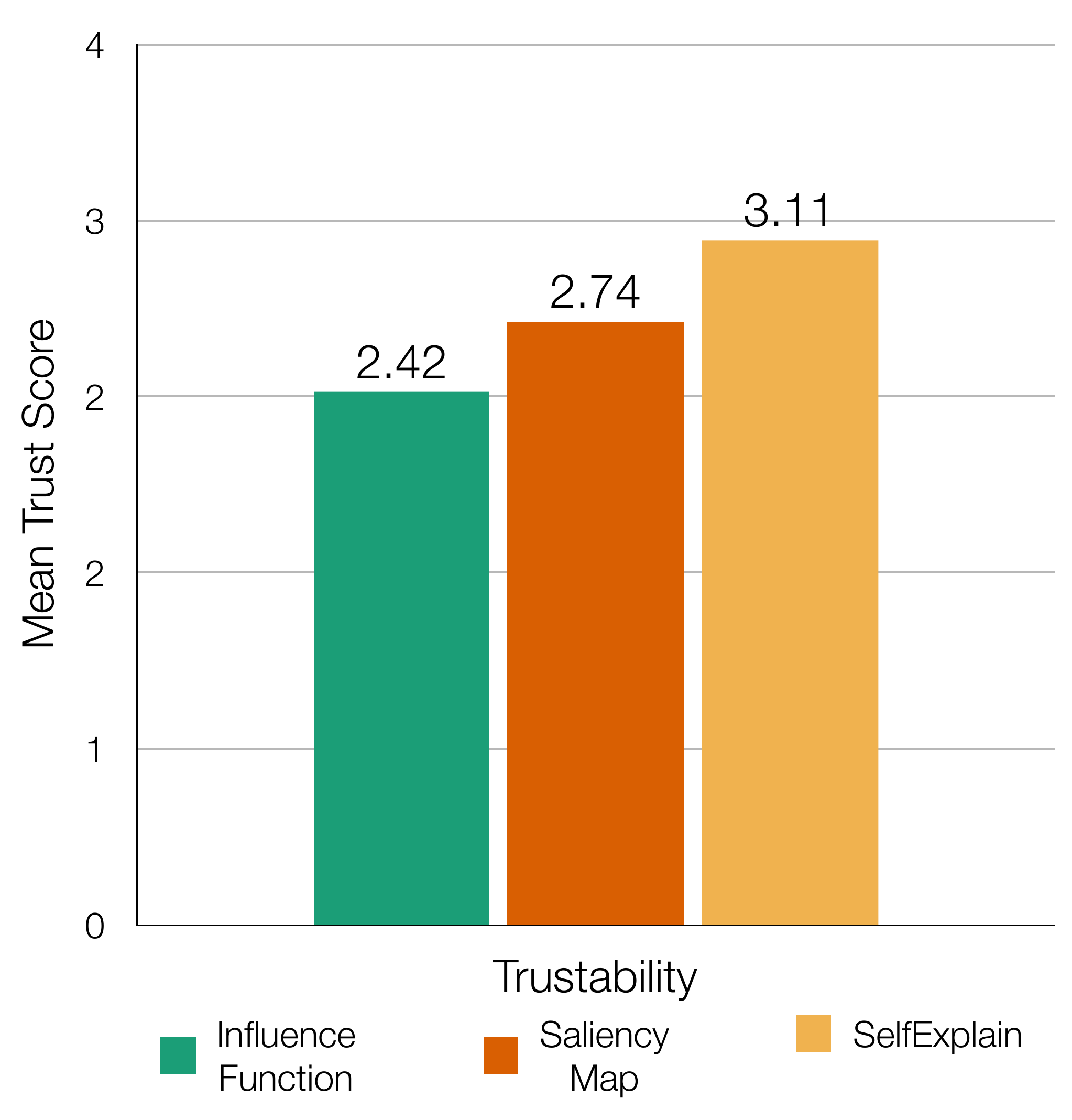}}
    \caption{\emph{Mean trust score} of \ourmodel~against baselines. The vertical axis show mean trust labeled on 1-5 likert scale. Humans judge \ourmodel~ explanations improve trust in model predictions.}
    \label{fig:trust_interpret}
\end{figure}

\paragraph{Understandability:} An essential criterion for  transparency in an AI system is the ability of a user to \emph{understand} model explanations  \citep{DoshiVelez2017AccountabilityOA}. 
Our understandability metric evaluates whether a human judge can understand the explanations presented by the model, which would equip a non-expert to verify the model predictions. Human judges were presented (i) the input, (ii) gold label, (iii) sentiment label prediction, and (iv) explanations from different methods (baselines, and \ourmodel), and were asked to select the explanation that they perceived to be more understandable. 
Figure \ref{fig:eval_interpret} (right) shows that
\ourmodel~ achieves 29\% improvement over the best-performing baseline in terms of understandability of the model explanation.

\paragraph{Trustability:} In addition to plausibility, we also evaluate user \emph{trust}  of the explanations \citep{singh2018hierarchical,Jin2020Towards}. To evaluate user trust, We follow the same experimental setup as \citet{singh2018hierarchical} and \citet{Jin2020Towards} to compute the \emph{mean trust score}. For each data sample, subjects were shown explanations and the model prediction from the three interpretability methods and were asked to rate on a Likert scale of 1--5 based on how much trust did each of the model explanations instill.
Figure \ref{fig:trust_interpret} shows the mean-trust score of \ourmodel~ in comparison to the baselines.  
We observe from the results that concept-based explanations are perceived more trustworthy for humans. 
\section{Analysis} 

Table \ref{tab:qual_examples} shows example interpretations by \ourmodel; we show some additional analysis of explanations from \ourmodel\footnote{additional analysis in appendix due to space constraints} in this section. \\

\noindent
\textbf{Does \ourmodel's explanation help predict model behavior?} In this setup, humans are presented with an explanation and an input, and
must correctly predict the model’s output \citep{doshi2017towards,Lertvittayakumjorn2019HumangroundedEO,hase-bansal-2020-evaluating}. 
We randomly selected 16 samples  
spanning equal number of true positives, true negatives, false positives and false negatives from the dev set. 
Three human judges were tasked to predict the model decision with and without the presence of model explanation. We observe that when users were presented with the explanation, their ability to predict model decision improved by an average of 22\%, showing that with \ourmodel's explanations, humans could better understand model's behavior.  \\

\paragraph{Performance Analysis:}
In \GIL, we study the performance trade-off of varying the number of retrieved influential concepts $K$.
From a performance perspective, there is only marginal drop in moving from the base model to \ourmodel~model with both \GIL~and \LIL~(shown in Table \ref{tab:gil_k_analysis}). From our experiments with human judges, we found that for sentence level classification tasks $K=5$ is preferable for a balance of performance and the ease of interpretability.

\begin{table}[!ht]
\centering
\begin{tabular}{@{}ccc@{}}
\toprule
\GIL~top-$K$ & steps/sec & memory \\ \midrule
base & 2.74 & 1$x$ \\
$K$=5* & 2.50 & 1.03$x$ \\
$K$=100 & 2.48 & 1.04$x$ \\
$K$=1000 & 2.20 & 1.07$x$ \\ \bottomrule
\end{tabular}
\caption{Effect of $K$ from \GIL. We use \ourmodel-\xlnet~on SST-2 for this analysis. *$K$=1/5/10 did not show considerable difference among them}
\label{tab:gil_k_analysis}
\end{table}

\paragraph{\LIL-\GIL-Linear layer agreement:} To understand whether our explanations lead to predicting the same label as the model's prediction, we analyze whether the final logits activations on the \GIL~and \LIL~layers agree with the linear layer activations. Towards this, we compute an agreement between label distributions from \GIL~and \LIL~layers to the distribution of the linear layer.  Our \LIL-\emph{linear} F1 is 96.6\%, \GIL-\emph{linear} F1 100\% and \GIL-\LIL-\emph{linear} F1 agreement is 96.6\% for \ourmodel-\xlnet~ on the SST-2 dataset. We observe that the agreement rates between the \GIL~, \LIL~ and the linear layer are very high, validating that \ourmodel's layers agree on the same model classification prediction, showing that \GIL~and \LIL~concepts lead to same predictions.

\paragraph{Are \LIL~concepts relevant?} For this analysis, we randomly selected 50 samples from SST2 dev set and removed the top most salient phrases ranked by \LIL. Annotators were asked to predict the label without the most relevant local concept and the accuracy dropped by 7\%. We also computed the \ourmodel-\xlnet~classifier's accuracy on the same input and the accuracy dropped by ${\sim}14\%$.\footnote{Statistically significant by Wilson interval test.} This suggests that \LIL~ captures  relevant local concepts.\footnote{Samples from this experiment are shown in \S\ref{subsec:relevance_examples}.} 

\begin{table*}[ht]
\centering
\resizebox{\textwidth}{!}{%
\begin{tabular}{@{}lll@{}}
\toprule
Input & \begin{tabular}[c]{@{}l@{}} Top \LIL~interpretations \end{tabular} & \begin{tabular}[c]{@{}l@{}}Top \GIL~interpretations \end{tabular} \\ \midrule
\begin{tabular}[c]{@{}l@{}}it 's a very charming \\ and often affecting journey\end{tabular} & \begin{tabular}[c]{@{}l@{}}often affecting, \\ very charming\end{tabular} & \begin{tabular}[c]{@{}l@{}}scenes of cinematic perfection that steal your heart away, \\ submerged, that extravagantly\end{tabular} \\ \midrule
\begin{tabular}[c]{@{}l@{}}it  ' s a charming and often \\ affecting journey of people\end{tabular} & \begin{tabular}[c]{@{}l@{}}of people, \\ charming and often affecting\end{tabular} & \begin{tabular}[c]{@{}l@{}}scenes of cinematic perfection that steal your heart away, \\ submerged,  that extravagantly\end{tabular} \\ \bottomrule
\end{tabular}%
}
\caption{Sample (from SST-2) of an input perturbation lead to different local concepts, but global concepts remain stable.}
\label{tab:similarity_example}
\end{table*}

\paragraph{Stability: do similar examples have similar explanations?} \citet{melis2018towards} argue that a crucial property that interpretable models need to address is \emph{stability}, where the model should be robust enough that a minimal change in the input should not lead to drastic changes in the observed interpretations. We qualitatively analyze this by measuring the overlap of \ourmodel's extracted concepts for similar examples. 
Table~\ref{tab:similarity_example} shows a representative example in which minor variations in the input lead to differently ranked local phrases, but their global influential concepts remain stable. 

\section{Related Work} 

\paragraph{Post-hoc Interpretation Methods:}
Predominant based methods for post-hoc interpretability in NLP use gradient based methods \citep{Simonyan2014DeepIC,sundararajan2017axiomatic,smilkov2017smoothgrad}. 
Other post-hoc interpretability methods such as \citet{singh2018hierarchical} and \citet{Jin2020Towards} decompose relevant and irrelevant aspects from hidden states and obtain a relevance score.
While the methods above focus on local interpretability, works such as \citet{han2020explaining} aim to retrieve influential training samples for global interpretations. Global interpretability methods are useful not only to facilitate explainability, but also to detect and mitigate artifacts in data \citep{pezeshkpour2021combining,han2021influence-tuning}.

\paragraph{Inherently Intepretable Models:} Heat maps based on attention \citep{bahdanau2014neural} are one of the commonly used interpretability tools for many downstream tasks such as machine translation \citep{luong-etal-2015-effective}, summarization \citep{rush-etal-2015-neural} and reading comprehension \citet{NIPS2015_5945}.
Another recent line of work explores collecting \emph{rationales} \citep{lei-etal-2016-rationalizing} through expert annotations \citep{zaidan2008modeling}. Notable work in collecting external rationales include Cos-E \citep{rajani2019salesforceexplain}, e-SNLI \citep{Camburu2018eSNLI} and recently, Eraser benchmark \citep{deyoung-etal-2020-eraser}. 
Alternative lines of work in this class of models include \citet{card2019deep} that relies on interpreting a given sample as a weighted sum of the training samples while \citet{croce2019auditing} identifies influential training samples using a kernel-based transformation function. \citet{jiang-bansal-2019-self} produce interpretations of a given sample through modular architectures, where model decisions are explained through outputs of intermediate modules.
A class of inherently interpretable classifiers explain model predictions locally using human-understandable high-level \emph{concepts} such as prototypes \citep{melis2018towards,ChenEtAl2019} and interpretable classes \citep{koh2020concept,yeh2020completenessaware}. They were recently proposed for computer vision applications, but despite their promise have not yet been adopted in NLP.
\ourmodel~is similar in spirit to \citet{melis2018towards} but additionally provides explanations via training data concepts for neural text classification tasks. 
\section{Conclusion}
In this paper, 
we propose \ourmodel, a novel self-explaining framework that enables explanations through higher-level concepts, improving from low-level word attributions.
\ourmodel~ provides both local explanations (via relevance of each input concept) and global explanations (through influential concepts from the training data) in a single framework via two novel modules (\LIL~ and \GIL), and trainable end-to-end.
Through human evaluation, we show that our interpreted model outputs are perceived  as more trustworthy, understandable, and adequate for explaining model decisions compared to previous approaches to explainability. 

This opens an exciting research direction for building inherently interpretable models for text classification. Future work will extend the framework to other tasks and to longer contexts, beyond single input sentence. We will also explore  additional approaches to extract target local and global concepts, including abstract syntactic, semantic, and pragmatic linguistic features. Finally, we will study what is the right level of abstraction for generating explanations for each of these tasks in a human-friendly way. 

\section*{Acknowledgements}

This material is based upon work funded by the DARPA CMO under Contract No.~HR001120C0124, and by the United States Department of Energy (DOE) National Nuclear Security Administration (NNSA) Office of Defense Nuclear Nonproliferation Research and Development (DNN R\&D) Next-Generation AI research portfolio. The views and opinions of authors expressed herein do not necessarily state or reflect those of the United States Government or any agency thereof.

\bibliography{naacl}
\bibliographystyle{acl_natbib}
\clearpage
\appendix
\section{Appendix}

\subsection{Additional Analysis}
\label{subsec:appendix_analysis}

\begin{table*}[ht]
\centering
\resizebox{\textwidth}{!}{%
\begin{tabular}{@{}lll@{}}
\toprule
Input & \begin{tabular}[c]{@{}l@{}} Top \LIL~interpretations \end{tabular} & \begin{tabular}[c]{@{}l@{}}Top \GIL~interpretations \end{tabular} \\ \midrule
\begin{tabular}[c]{@{}l@{}}it 's a very charming \\ and often affecting journey\end{tabular} & \begin{tabular}[c]{@{}l@{}}often affecting, \\ very charming\end{tabular} & \begin{tabular}[c]{@{}l@{}}scenes of cinematic perfection that steal your heart away, \\ submerged, that extravagantly\end{tabular} \\ \midrule
\begin{tabular}[c]{@{}l@{}}it  ' s a charming and often \\ affecting journey of people\end{tabular} & \begin{tabular}[c]{@{}l@{}}of people, \\ charming and often affecting\end{tabular} & \begin{tabular}[c]{@{}l@{}}scenes of cinematic perfection that steal your heart away, \\ submerged,  that extravagantly\end{tabular} \\ \bottomrule
\end{tabular}%
}
\caption{Sample (from SST-2) of an input perturbation lead to different local concepts, but global concepts remain stable.}
\label{tab:similarity_example}
\end{table*}

\paragraph{Stability: do similar examples have similar explanations?} \citet{melis2018towards} argue that a crucial property that interpretable models need to address is \emph{stability}, where the model should be robust enough that a minimal change in the input should not lead to drastic changes in the observed interpretations. We qualitatively analyze this by measuring the overlap of \ourmodel's extracted concepts for similar examples. 
Table~\ref{tab:similarity_example} shows a representative example in which minor variations in the input lead to differently ranked local phrases, but their global influential concepts remain stable.

\subsection{Qualitative Examples}
\label{subsec:qual_examples}

Table \ref{tab:qual_examples_appendix} shows some qualitative examples from our best performing SST-2 model. 

\begin{table*}[]
\centering
\resizebox{\textwidth}{!}{%
\begin{tabular}{@{}lll@{}}
\toprule
Input Sentence & Explanation from Input & Explanation from Training Data \\ \midrule
\begin{tabular}[c]{@{}l@{}}offers much to enjoy ... \\ and a lot to mull over in  terms of love ,\\  loyalty and the nature of staying friends .\end{tabular} & ['much to enjoy', 'to enjoy', 'to mull over'] & \begin{tabular}[c]{@{}l@{}}['feel like you ate a reeses \\\ without the peanut butter']\end{tabular} \\
\begin{tabular}[c]{@{}l@{}}puts a human face on a land most \\ westerners are unfamiliar with .\end{tabular} & \begin{tabular}[c]{@{}l@{}}['put s a human face on a land most \\  westerners are unfamiliar with',\\  'a human face']\end{tabular} & ['dazzle and delight us'] \\
nervous breakdowns are not entertaining . & ['n erv ous breakdown s', 'are not entertaining'] & ['mesmerizing portrait'] \\
too slow , too long and too little happens . & ['too long', 'too little happens', 'too little'] & \begin{tabular}[c]{@{}l@{}}['his reserved but existential poignancy', \\ 'very moving and revelatory footnote']\end{tabular} \\
very bad . & ['very bad'] & \begin{tabular}[c]{@{}l@{}}['held my interest precisely',\\  'intriguing , observant', \\ 'held my interest']\end{tabular} \\
\begin{tabular}[c]{@{}l@{}}it haunts , horrifies , startles and fascinates ;\\  it is impossible to look away .\end{tabular} & \begin{tabular}[c]{@{}l@{}}['to look away', 'look away',\\  'it haun ts , horr ifies , start les and fasc inates']\end{tabular} & \begin{tabular}[c]{@{}l@{}}['feel like you ate a reeses \\ without the peanut butter']\end{tabular} \\
it treats women like idiots . & ['treats women like idiots', 'like idiots'] & \begin{tabular}[c]{@{}l@{}}[ 'neither amusing \\ nor dramatic enough \\ to sustain interest']\end{tabular} \\
\begin{tabular}[c]{@{}l@{}}the director knows how to apply textural gloss , \\ but his portrait of sex-as-war is strictly sitcom .\end{tabular} & \begin{tabular}[c]{@{}l@{}}['the director', \\  'his portrait of sex - as - war']\end{tabular} & \begin{tabular}[c]{@{}l@{}}[ 'absurd plot twists' ,\\  'idiotic court maneuvers \\    and stupid characters']\end{tabular} \\
too much of the humor falls flat . & \begin{tabular}[c]{@{}l@{}}['too much of the humor', \\ 'too much', 'falls flat']\end{tabular} & ['infuriating'] \\
\begin{tabular}[c]{@{}l@{}}the jabs it employs are short , \\ carefully placed and dead-center .\end{tabular} & \begin{tabular}[c]{@{}l@{}}['it employs',\\  'carefully placed', 'the j abs it employs']\end{tabular} & ['with terrific flair'] \\
\begin{tabular}[c]{@{}l@{}}the words , ` frankly , my dear , \\ i do n't give a damn ,\\ have never been more appropriate .\end{tabular} & ["do n 't give a damn"] & ['spiteful idiots'] \\
\begin{tabular}[c]{@{}l@{}}one of the best films of the year with its \\ exploration of the obstacles \\ to happiness faced by five contemporary \\ individuals ... a psychological masterpiece .\end{tabular} & \begin{tabular}[c]{@{}l@{}}['of the best films of the year', \\ 'of the year', 'the year']\end{tabular} & ['bang'] \\
\begin{tabular}[c]{@{}l@{}}my wife is an actress is an utterly \\ charming french comedy that feels so \\ american in sensibility and style it 's\\ virtually its own hollywood remake .\end{tabular} & \begin{tabular}[c]{@{}l@{}}['an utterly charming french comedy', \\ 'utterly charming', 'my wife']\end{tabular} & ['all surface psychodramatics'] \\ \bottomrule
\end{tabular}%
}
\caption{Samples from \ourmodel's interpreted output. }
\label{tab:qual_examples_appendix}
\end{table*}

\subsection{Relevant Concept Removal}
\label{subsec:relevance_examples} 

Table \ref{tab:lil_examples} shows us the samples where the model flipped the label after the most relevant local concept was removed. In this table, we show the original input, the perturbed input after removing the most relevant local concept, and the corresponding model predictions. 

\begin{table*}[]
\centering
\resizebox{\textwidth}{!}{%
\begin{tabular}{@{}llll@{}}
\toprule
Original Input & Perturbed Input & \begin{tabular}[c]{@{}l@{}}Original \\ Prediction\end{tabular} & \begin{tabular}[c]{@{}l@{}}Perturbed\\ Prediction\end{tabular} \\ \midrule
unflinchingly bleak and desperate & unflinch \_\_\_\_\_\_\_\_\_\_\_\_\_\_\_\_ & negative & positive \\
\begin{tabular}[c]{@{}l@{}}the acting , costumes , music , \\ cinematography and sound are all \\ astounding given the production 's\\  austere locales .\end{tabular} & \begin{tabular}[c]{@{}l@{}}\_\_\_\_\_\_\_\_ , costumes , music , cinematography \\ and sound are all astounding given the\\  production 's austere locales .\end{tabular} & positive & negative \\
\begin{tabular}[c]{@{}l@{}}we root for ( clara and paul ) , \\ even like them , \\ though perhaps it 's an emotion\\  closer to pity .\end{tabular} & \begin{tabular}[c]{@{}l@{}}we root for ( clara and paul ) ,\_\_\_\_\_\_\_\_\_\_\_ ,\\  though perhaps it 's an emotion closer to pity .\end{tabular} & positive & negative \\
\begin{tabular}[c]{@{}l@{}}the emotions are raw and will strike\\ a nerve with anyone who 's ever \\ had family trauma .\end{tabular} & \begin{tabular}[c]{@{}l@{}}\_\_\_\_\_\_\_\_\_\_ are raw and will strike a\\  nerve with anyone who 's ever had family trauma .\end{tabular} & positive & negative \\
holden caulfield did it better . & holden caulfield \_\_\_\_\_\_\_\_\_\_ . & negative & positive \\
\begin{tabular}[c]{@{}l@{}}it 's an offbeat treat that pokes fun at the \\ democratic exercise while also \\ examining its significance for those who take part .\end{tabular} & \begin{tabular}[c]{@{}l@{}}it 's an offbeat treat that pokes \\ fun at the democratic exercise\\  while also examining \_\_\_\_\_\_\_\_\_ for \\ those who take part .\end{tabular} & positive & negative \\
\begin{tabular}[c]{@{}l@{}}as surreal as a dream and as detailed as a \\ photograph , as visually dexterous \\ as it is at times imaginatively overwhelming .\end{tabular} & \begin{tabular}[c]{@{}l@{}}\_\_\_\_\_\_\_\_\_\_\_\_\_\_\_ and as detailed as a photograph ,\\  as visually dexterous as it is at times\\  imaginatively overwhelming .\end{tabular} & positive & negative \\
\begin{tabular}[c]{@{}l@{}}holm ... embodies the character with\\  an effortlessly regal charisma .\end{tabular} & \begin{tabular}[c]{@{}l@{}}holm ... embodies the\\  character with \_\_\_\_\_\_\_\_\_\_\_\_\end{tabular} & positive & negative \\
\begin{tabular}[c]{@{}l@{}}it 's hampered by a lifetime-channel \\ kind of plot and a lead actress who is out of her depth .\end{tabular} & \begin{tabular}[c]{@{}l@{}}it 's hampered by a \\ lifetime-channel kind of \\ plot and a lead actress\\  who is \_\_\_\_\_\_\_\_\_\_\_\_ .\end{tabular} & negative & negative \\ \bottomrule
\end{tabular}%
}
\caption{Samples where the model predictions flipped after removing the most relevant local concept. }
\label{tab:lil_examples}
\end{table*}

\end{document}